\documentclass[conference]{IEEEtran}
\IEEEoverridecommandlockouts
\usepackage{cite}
\usepackage{amsmath,amssymb,amsfonts}
\usepackage{algorithmic}
\usepackage{graphicx}
\usepackage{textcomp}
\usepackage{xcolor}
\def\BibTeX{{\rm B\kern-.05em{\sc i\kern-.025em b}\kern-.08em
    T\kern-.1667em\lower.7ex\hbox{E}\kern-.125emX}}
\begin{document}

\title{Ensemble Method for Cluster Number Determination and Algorithm Selection in Unsupervised Learning*
\thanks{*This is a preprint.}}

\author{\IEEEauthorblockN{Antoine E. Zambelli}
}

\maketitle

\begin{abstract}
Unsupervised learning, and more specifically clustering, suffers from the need for expertise in the field to be of use. Researchers must make careful and informed decisions on which algorithm to use with which set of hyperparameters for a given dataset. Additionally, researchers may need to determine the number of clusters in the dataset, which is unfortunately itself an input to most clustering algorithms. All of this before embarking on their actual subject matter work. After quantifying the impact of algorithm and hyperparameter selection, we propose an ensemble clustering framework which can be leveraged with minimal input. It can be used to determine both the number of clusters in the dataset and a suitable choice of algorithm to use for a given dataset. A code library is included in the Conclusion for ease of integration.
\end{abstract}


\section{Introduction}
\label{sec:1}


Unsupervised learning - one of the main branches of machine learning - is the study of previously unlabeled datasets. With the hope of gaining new insights into their data and its structure, researchers can attempt to group or segment their data based on similarities between the data points in an exercise called clustering.

Clustering problems have been studied in-depth, with applications in the fields of Computational Biology, Operations Research and Social Sciences~\cite{mcguirl,song,caoli} to name a few. Quite a few algorithms are available for use - including in popular open-source libraries~\cite{scikit-learn,fastcluster}. However, these problems have often required human intervention on the part of the researcher. Naturally, this severely limits automation and is prone to human-error.

The need for researcher input is due mainly to two problems: determining the number of clusters in the dataset, and choosing an algorithm to cluster with. Both can produce highly inaccurate results if poorly selected.

\subsection{The Landscape}

The canonical approach to determining the number of clusters in a dataset is to take an ``elbow method" approach. We detail this in Section~\ref{sec:2}. In short, attempt to cluster the dataset with different numbers of clusters and compare the outputs - looking for an elbow in the curve of results. However, this has inherent shortcomings. A researcher must still choose which algorithm and which set of hyperparameters to use.

While Hierarchical Clustering Algorithms (HCA) can be used to automatically find the number of clusters with reasonable accuracy in some cases~\cite{zambelli}, using this method means we miss out on using the many other algorithms that have been developed - even if they would be better suited to our dataset.

A large number of sophisticated methods have been explored as well~\cite{survey}. One of the more famous approaches was developed by Monti, et. al.~\cite{monti} - called Consensus Clustering. Many of these methods (including Monti) suffer from a high level of complexity and abstraction - often based on the idea of partitioning the data~\cite{ensemble,ensemble2}. Essentially they attempt to cluster many different subsets of the data under different cluster numbers and then select the most stable.

Apart from usability stemming from their complexity, many of these methods can get computationally intensive (including memory requirements). Lastly, as noted in~\cite{limitations}, they do not generally perform well when it comes to estimating the number of clusters. As a final note, these approaches also suffer from the need for researcher input as to the choice of algorithm and parameters. We consider it prudent, therefore, to explore the topic further.

In this paper, we propose an ensemble approach to answering the following questions: How many clusters are in the dataset, and which algorithm-hyperparameter choice is best for this data? Our approach outputs the number of clusters, as well as both a model choice and a set of hyperparameters to use with the model. Note that we do not define ensemble in the sense of a collection of partitions, but rather separate algorithms, as seen in~\cite{robust}.

Luckily, several of the graph-based methods mentioned above - including Consensus Clustering and its improvements - are actually complementary to our proposed method, and we see no reason why they could not be combined, albeit with a bit of work. Consensus Clustering takes as input a model and its hyperparameters, but has no framework for choosing a suitable model. Likewise, this first discovery we are presenting does not account for any data partitions.

While we leave the task of truly combining our approach with existing methods to future work, we present evidence of the benefits of accounting for model and hyperparameter choices in Monti's Consensus Clustering, and we invite the reader to consider the possibilities as they read through our work. In Section~\ref{sec:2}, we will look at the problem in more detail, exploring the three main areas that must be addressed. In Section~\ref{sec:3} we will define our algorithm or workflow. We will then discuss results in Section~\ref{sec:4}, present typical usage in Section~\ref{sec:5}, and finally summarize our approach and findings in Section~\ref{sec:6}.

\section{The Problem}
\label{sec:2}

As mentioned, the central issue we are facing is that of finding the number of clusters in our dataset. While we have previously stated that the choice of algorithm and hyperparameter set is important, we found it was often overlooked in the literature - being taken as a given or covered at a high level~\cite{review}. So, we set out to compute some baseline performance changes that choosing the algorithm and hyperparameter set can have. We found that, in fact, this choice had a very large impact on predicting the number of clusters in a dataset - and were ourselves quite startled by the magnitude of these variations.

Perhaps the most common approach to determining the number of clusters is to use the elbow method~\cite{statbook, elbow}. This involves making many attempts at clustering, and then picking the one that seems to fit best. More directly, the workflow is:
\begin{enumerate}
    \item Choose a clustering algorithm and parameter set.
    \item Cluster the data into $n$ clusters for $n \in \{ 2, \cdots, N \}$.
    \item For each $N-1$ attempts, compute a metric (ex: $BIC$).
    \item Find the elbow in the curve of metric values, the x-axis value is our number of clusters.
\end{enumerate}
(note that automatically finding the elbow can be done in several ways such as the minimum absolute second derivative, the point of best linear fit, or as we will use here the triangle method).

Intuitively, we can view this method through the lens of Information Theory. Namely, the curve represents the amount of information explained as we increase the number of clusters. As we begin getting diminishing returns, we say that we are no longer explaining the data, and have too many clusters (hence the cutoff at the elbow).

Our approach will rest on this method and principle, but will tackle its three weaknesses - the choices that were implicitly made: algorithm, hyperparameters and metric. Many of the difficulties are discussed at a high-level in~\cite{review}. We are essentially tackling Step 2 (and partially Step 1) in Fig 1 in their Conclusion.

For now, let's quantify exactly how important these choices are. Using the elbow method as a baseline, we'll compute some accuracy statistics on 100 randomly-generated 3-cluster datasets, detailed in Section~\ref{sec:21}. We define accuracy as correctly determining the number of clusters we have (in this case 3). For each algorithm, we look at performance across a broad range of hyperparameters and metrics - detailed in Section~\ref{sec:32}.

\subsection{Simulated Data}
\label{sec:21}

Our data consists of 100 2-dimensional 3-cluster collections of 30,000 points each. The data was drawn from a standard normal distribution, with cluster centers randomly places between $(-5,-5)$ and $(5,5)$. Specifically, the data was constructed using the scikit-learn function \verb|make_blobs|:

\begin{verbatim}
X, y = make_blobs(
    n_samples=30000,
    centers=3,
    n_features=2,
    center_box=(-5, 5),
    random_state=seed
)
\end{verbatim}

\subsection{Choice of Algorithm}

The first step many researchers will take to successfully cluster their data is to choose a clustering algorithm (step 1 in the traditional workflow above). A variety of inherently distinct algorithms exist, from Spectral methods to HCA and DBSCAN. The issue at this step is that different algorithms can be better suited to different datasets~\cite{gordon}, and this can be very difficult to determine ahead of time. While there are some generally accepted behaviors - ie, Spectral clustering works well on non-convex datasets~\cite{scikit-learn} - we can also see this experimentally.

For each algorithm, hyperparameter and metric choice, we compute the accuracy of our predicted number of clusters on the 100 datasets, giving us $M$ accuracy readings. For example, for 1 algorithm with 2 possible metrics and 3 hyperparameter values, we would have 6 accuracy readings for the clustering algorithm. If the choice of algorithm did not matter, then we would get the same statistics across different algorithms. The table below shows the statistics for those $M$ readings for K-means, Gaussian Mixture Model (GMM), HCA and Spectral:

\begin{table}[h!]
	\caption{Stats for accuracy readings per algorithm.}
	\centering\begin{tabular}{l|rrrr}
		
		Stat &  K-means & GMM &  HCA &  Spectral \\
		
		mean &  61.78 &  66.47 &  28.75 &  33.05 \\
		std &  27.52 &  33.29 &  16.09 &  24.00 \\
		min &  8.00 &  7.00 &  10.00 &  6.00 \\
		max &  85.00 &  91.00 &  78.00 &  72.00 \\
		
	\end{tabular}
\end{table}

As we can see, different algorithms obtain rather different results (whether the mean or max performance). Particularly interesting is the performance of the Spectral and HCA algorithms, which are clearly ill-suited to our datasets. On the other hand, we can see that there is some combination of metrics and hyperparameters for which GMM does quite well with a 91\% accuracy. Its standard deviation is quite high though, suggesting that different metrics and hyperparameters can lead to quite different results.

\subsection{Choice of Hyperparameters}

The second choice traditionally faced by researchers is hyperparameters (implicitly contained in step 2 of the traditional workflow). Generally referred to as hyperparameter tuning, this can greatly improve the performance of a model. Let's examine the effect of hyperparameters in our experimental setup from Section 2.1. Let's hold the choice of metric constant (selecting inertia $I$ - more on this in Section 2.4). This gives us:

\begin{table}[h!]
	\caption{Stats for accuracy readings per algorithm-metric, across hyperparameters.}
	\centering\begin{tabular}{l|rrrr}
		
		Stat &  $\text{K-means}_{I}$ & $\text{GMM}_{I}$ &  $\text{HCA}_{I}$ &  $\text{Spectral}_{I}$ \\
		
		mean &  77.63 &  85.39 &  41.25 &  51.33 \\
		std &  4.00 &  1.14 &  28.09 &  20.96 \\
		min &  71.00 &  83.00 &  10.00 &  10.00 \\
		max &  84.00 &  87.00 &  78.00 &  72.00 \\
		
	\end{tabular}
\end{table}

Table 2 shows us that the choice of hyperparameters, independent of other choices, can lead to large differences in performance. For instance, a judicious choice of hyperparameters in our K-means algorithm can lead to an 18\% increase in performance. Similarly, a poor choice for Spectral leads to a staggering 86\% drop. Unfortunately, we have no way of knowing ahead of time which parameter selection will yield the best results in a clustering problem (given an absence of ground truth).

Further, hyperparameters explain some of the variation we saw in Table 1 but not all. A quick look at the various statistics shows us we are missing another piece: the minimum performance seen by the K-means algorithm is now 71\% instead of the rather shocking 8\%. This indicates one more component to the problem.

\subsection{Choice of Metric}

Finally, we arrive at the last choice we have to make - which metric to use (step 3 in the aforementioned workflow). Previous work by the author showed that in the case of HCA different metrics performed differently~\cite{zambelli}, but there hasn't been much work on this topic in general. However, we can once again examine this experimentally. In the same experimental setup as we used above, let's examine the performance of algorithms for a fixed selection of hyperparameters across different metrics.

Every algorithm used the Inertia and Silhouette Score metrics~\cite{scikit-learn}. HCA also used the Maximum Difference and Elbow metrics from~\cite{zambelli}. K-means and GMM also used the AIC and BIC metrics~\cite{scikit-learn}. For the table of results, we take the first set of hyperarameters for that algorithm, denoted by the superscript $0$.

\begin{table}[h!]
	\caption{Stats for accuracy readings per algorithm-hyperparameter, across metrics.}
	\centering\begin{tabular}{l|rrrr}
		
		Stat &  $\text{K-means}^{0}$ & $\text{GMM}^{0}$ &  $\text{HCA}^{0}$ &  $\text{Spectral}^{0}$ \\
		
		mean &  61.75 &  64.50 &  24.75 &  16.50 \\
		std &  33.17 &  37.70 &  7.59 &  9.19 \\
		min &  12.00 &  8.00 &  17.00 &  10.0 \\
		max &  79.00 &  85.00 &  35.00 &  23.00 \\
		
	\end{tabular}
\end{table}

Once again, we can see variations in performance within each algorithm-hyperparameter choice, here based solely on the choice of metrics. It turns out that much like we saw in Table 1 with HCA and Spectral, there is a weak element: the Silhouette Score.

Though not obvious from this table, further investigation shows that it has an average accuracy across all algorithms and hyperparameters of only 12\%. By cutting the number of algorithm-hyperparmeter combinations we've amplified its effect, and are seeing it drag performance down across the board - something we couldn't know ahead of time.

The problem is now apparent: we need to find a way to filter out the many possibly bad choices in algorithms, hyperparameters and metrics if we are to find the number of clusters in a dataset with reasonable accuracy.

\section{Ensemble Approach}
\label{sec:3}
As we saw in the previous section, we know there are some winning combinations of algorithm-hyperparameter-metric, but we must now figure out how to find them ahead of time. At a high level, our approach will use a fairly simple ensemble method. We will cluster the dataset using all the combinations we can think of and select our predicted number of clusters from all the results combined.

This section's structure will follow the workflow of our algorithm, split into subsections 3.2 through 3.4. While we have found a preferred approach on our test dataset, we present several alternatives to each step of the workflow. First, however, the reader must endure some exposition of notation to be used throughout.

Suppose we work with an ensemble of algorithms, denoted by the set
\begin{equation}
    \mathcal{A} = \left\{A_0,\cdots,A_N\right\}.
\end{equation}
These would be the clustering algorithms such as K-means, GMM, etc.

Each algorithm can have a set of hyperparameters associated with it. Let that be written as
\begin{equation}
    \mathcal{H}^j = \left\{h_0^j,\cdots,h_{H_j}^j\right\}, \quad\forall j\in[0,N]
\end{equation}
where $h_{i}^j$ denotes the $i$th hyperparameter selection for the $j$th algorithm. Note that different algorithms can have different numbers of hyperparameter selections (denoted by $H_j$).

Further, each $h_i^j$ can be comprised of several elements. For instance for GMM, $h_0^{1}$ could be the covariance type and regularization parameter: $(\textsc{diagonal}, 10^{-6})$, and $h_1^{1} = (\textsc{diagonal}, 10^{-5})$. If it helps, think of each $h_{i}^j$ as a set of \textsc{kwargs} passed into a model object.

Lastly, let's write the set of metrics (inertia, AIC, etc) used as
\begin{equation}
    \mathcal{M}^j =  \left\{m_0^j,\cdots,m_{M_j}^j\right\},\quad\forall j\in[0,N]
\end{equation}
where $m_i^j$ is the $i$th metric used to evaluate the $j$th algorithm.

Now, given $\mathcal{A}$, $\mathcal{H}^j$ and $\mathcal{M}^j$, we have $\forall j\in [0,N] $
\begin{align}
    \mathcal{P}^j &= \mathcal{H}^j\times\mathcal{M}^j\\
    &= \left\{\left(h_0^j,m_0^j\right),\cdots,\left(h_{H_j}^j,m_{M_j}^j\right)\right\}
\end{align}
and
\begin{equation}
    \mathcal{C}^j = \left\{A_j(p) \ \middle| \ p\in\mathcal{P}^j\right\}.
\end{equation}
We admit the notation is somewhat opaque, but by constructing our actual test sets it should be illustrated nicely.

\subsection{Workflow}

Now that we have defined our $\mathcal{A}$, $\mathcal{H}$ and $\mathcal{M}$ sets - and obtained our $\mathcal{P}$ sets - we must compute the clusterings.

This is, quite simply, an exhaustive loop over all the elements of the respective $\mathcal{P}$ set, where we apply the elbow method as described in Section 2. $\forall j \in [0,N]$, and $\forall p \in \mathcal{P}^j$,
\begin{enumerate}
    \item We take an element $p\in\mathcal{P}^j$.
    \item We use the hyperparameter values from $p$ to compute clusterings for a range of cluster numbers.
    \item We then use the metric found in $p$ to get an elbow curve.
    \item We find the elbow in the curve.
\end{enumerate}
For each $p$, we have now found a suitable number of clusters for our data. More formally, we have just computed
\begin{equation}
    \mathcal{C}^j = \left\{ A^j(p)\ \middle|\ p\in\mathcal{P}^j \right\}.
\end{equation}
Note that $\mathcal{C}^j$ is simply a set of integers that map back to specific elements in $\mathcal{P}^j$. For K-means clustering a $3$-cluster dataset, we might get
\begin{equation}
    \mathcal{C}^0 = \left\{ 3,3,2,4,3,\cdots,3 \right\}
\end{equation}
We can then combine the results into a collection and find the number of clusters and best algorithm-hyperparameter selection from there:
\begin{enumerate}
    \item Construct the ensemble set $\mathcal{E}$ (we develop two approaches detailed in Section~\ref{sec:33}).
    \item Vote on the number of clusters (we develop three approaches detailed in Section~\ref{sec:34}).
\end{enumerate}

\subsection{Set Construction}
\label{sec:32}
The first step in the workflow is to construct our $\mathcal{A}$, $\mathcal{H}$, and $\mathcal{M}$ sets. Let's build some actual test sets to illustrate the structure. These will be used for computational results in Section 4, and might help clarify the notation in the meantime. These models are built from the \verb!scikit-learn! and \verb!fastcluster! libraries.

First, let's define the algorithms to look at:
\begin{equation}
    \mathcal{A} = \left\{ \text{K-means}, \text{GMM}, \text{HCA}, \text{Spectral} \right\}.
\end{equation}
These were selected based on their diverse natures. Ideally, one wants to choose a collection of algorithms that work well on different types of data. In this case, K-means is fast and reasonably accurate algorithm for convex datasets, HCA is fundamentally different and does not take cluster numbers as inputs, GMM is well-suited to data with a roughly Gaussian distribution and Spectral has been known to do well with non-convex datasets.

Now, the set of hyperparameters can get rather cumbersome to write out, but let's explicitly list those ranges for K-means:
\begin{equation}
    \verb!init! :\ \left\{ \verb!k-means++!,\ \verb!random! \right\}
\end{equation}
and
\begin{equation*}
    \verb!reassignment_ratio! :\quad\quad\quad\quad\quad\quad\quad\quad
\end{equation*}
\begin{equation}
    \quad\verb!np.geomspace(1e-4, 0.5, 8)!
\end{equation}
which is
\begin{verbatim}
[1.000e-04 3.376e-04 1.139e-03 3.848e-03
 1.299e-02 4.386e-02 1.480e-01 5.000e-01].
\end{verbatim}
This gives us the following set of hyperparameters for K-means:
\begin{align}
    \mathcal{H}^0 &= \left\{h_0^0,\cdots,h_{16}^0\right\}\\
    &= \left\{ \left(\mathtt{k}\text{-}\mathtt{means}\text{++}, \mathtt{1.000e}\text{-}\mathtt{04}\right), \right.\nonumber\\
    &\quad \left. \left(\mathtt{k}\text{-}\mathtt{means}\text{++}, \mathtt{3.376e}\text{-}\mathtt{04}\right),\right.\nonumber\\
    &\quad \left.\cdots\right.\nonumber\\
    &\quad \left. \left(\mathtt{random}, \mathtt{5.000e}\text{-}\mathtt{01}\right)\right\}.
\end{align}
For brevity, we present the remaining sets based on their base ranges:
\begin{verbatim}
GMM
covariance_type: [diag, tied, spherical],
reg_covar: np.geomspace(1e-8, 1e-2, 6)

HCA
method: [centroid, median, single, ward],
metric: [euclidean]

Spectral
affinity:
  [laplacian, precomputed, rbf, sigmoid],
metric: [cosine, l2, l1],
n_neighbors: [5, 20, 100],
gamma: [0.1, 1.0, 10.0]
\end{verbatim}
(where, for Spectral, \verb!metric! and \verb!n_neighbors! are only used for \verb!precomputed!, and \verb!gamma! is ignored for \verb!precomputed!).

In general, hyperparameters should be selected based on available information. If a researcher can somehow narrow the hyperparameter space through other knowledge, they should do so. In the absence of such information, as is our case here, we try to choose hyperparameter ranges that span the space.

Obviously, some of these parameters can take on an infinite number of values (and we have limited computing resources), but we find it judicious to choose a smaller number of values across orders of magnitude to obtain a representative sample of reasonable values.

Now that we have our $\mathcal{A}$ and $\mathcal{H}$ sets, we need $\mathcal{M}$. Per algorithm, we have:
\begin{verbatim}
K-means
[aic, bic, inertia, silhouette_score]

GMM
[aic, bic, inertia, silhouette_score]

HCA
[elbow, inertia, silhouette_score, max_diff]

Spectral
[inertia, silhouette_score]
\end{verbatim}

In other words, for K-means, we get:
\begin{align}
    \mathcal{P}^0 &= \left\{ \left(\mathtt{k}\text{-}\mathtt{means}\text{++}, \mathtt{1.000e}\text{-}\mathtt{04},\mathtt{aic} \right),\cdots \right.\nonumber\\
    &\left. \left(\mathtt{random}, \mathtt{5.000e}\text{-}\mathtt{01},\mathtt{silhouette} \right), \cdots \right\}.
\end{align}
What remains now is to construct our ensemble collection $\mathcal{E}$.

\subsection{Building the Ensemble}
\label{sec:33}
We present two approaches for building the ensemble set $\mathcal{E}$ in the following sections, and detail the rest of the workflow thereafter.

\subsubsection{Raw}

Given our $\mathcal{C}$ sets, the most natural way to construct our ensemble is simply to check every possible combination - essentially a cross product of our sets. This gives us our ensemble of values
\begin{align}
    \mathcal{E} &= \mathcal{C}^0\times\cdots\times \mathcal{C}^N\\
    &= \left\{A_0(p) \ \middle| \ p\in\mathcal{P}^0\right\} \times \cdots \times \left\{A_N(p) \ \middle| \ p\in\mathcal{P}^N\right\}\\
    &= \left\{A_j\left(p_i^j\right) \ \middle| \ j \in[0,N],\ i\in[0,H_jM_j]\right\}\\
    &= \left\{ \left( A_0 \left(p_0^0\right),\cdots,A_N \left(p_0^N\right) \right),\right.\nonumber\\
    &\quad\left.\cdots,\left( A_0 \left(p_{H_0M_0}^0\right),\cdots,A_N \left(p_{H_NM_N}^N\right) \right) \right\}.
\end{align}
Following the example from the previous section, $\mathcal{E}$ would be comprised of 4-tuples spanning all possible combinations. Each tuple would contain a ``guessed" cluster number given a specific algorithm-hyperparameter-metric choice.

If we structured $\mathcal{E}$ as a matrix, the first few rows might look like
\begin{verbatim}
E = [3 3 2 3]
    [3 3 2 4]
    [3 3 2 2]
    [3 3 2 5]
\end{verbatim}
where the first column corresponds to guesses from K-means, the second from GMM, then HCA and Spectral algorithms.

\subsubsection{Mode}

While the Raw approach detailed above is the simplest, we consider the fact that it ignores a potentially important point. The choice of metric, while critical to the workflow, is intrinsically an "elbow-method" parameter. This sets it apart from the choice of algorithm and hyperparameters, which would be necessary regardless of approach.

With this in mind, we consider another formulation which first takes the mode of the results across metrics. That is, for a given algorithm and hyperparameter configuration, we take as a result the most commonly guessed number of clusters across all metric choices. Define
\begin{align}
    \overline{\mathcal{C}}^j =& \left\{ Mode \left(\left\{ A_j(h_0^j, m) \mid m\in \mathcal{M}^j \right\}\right),\cdots, \right.\nonumber\\
    &\left. Mode \left(\left\{ A_j(h_{H_j}^j, m) \mid m\in \mathcal{M}^j \right\}\right)  \right\}\\
    =& \left\{ Mode \left(\left\{ A_j(h, m) \mid m\in \mathcal{M}^j \right\}\right) \mid h\in\mathcal{H}^j \right\}.
\end{align}

From this we can define our ensemble $\overline{\mathcal{E}}$ in the same way as $\mathcal{E}$,
\begin{equation}
    \overline{\mathcal{E}} = \overline{\mathcal{C}}^0\times\cdots\times \overline{\mathcal{C}}^N,
\end{equation}
which might give us an example matrix of
\begin{verbatim}
E = [3 3 3 3]
    [3 2 2 4]
    [3 3 2 2]
    [3 3 2 3]
\end{verbatim}
Here, our matrix will be smaller than in the Raw approach. Each column still corresponds to an algorithm, but each entry is now the mode of the guesses produced by a set of hyperparameters.

\subsection{Voting}
\label{sec:34}
Now, given our matrix $E$, there are a few ways to combine the results and vote on them. As a toy example, consider this result for a 3-cluster dataset:
\begin{verbatim}
E = [2 2 2 2]
    [2 2 2 2]
    [3 2 3 3]
    [3 3 2 3]
    [3 3 3 2]
\end{verbatim}

\subsubsection{Full}

The simplest approach would be to simply take the most common cluster number found in our ensemble. While straightforward, it doesn't allow us to capture any additional information, nor filter out any errors or biases in any way. Our toy example contains 11 $2$s and 9 $3$s, we would get an incorrect final result of $R=2$.

\subsubsection{Column-First}

Another naive approach would be to vote along algorithms, giving us 4 results, and then voting for the most common answer within those 4. One possible issue with this approach is the case where we have a few particularly ill-suited algorithms. Looking at the same example, we get
\begin{verbatim}
3 2 2 2
\end{verbatim}
and an incorrect final result of $R = 2$.

\subsubsection{Row-First}

Finally, we can look to capture what we are calling the cohesion between the results - essentially, favoring their agreement. By first looking at the individual rows of our example we would have
\begin{verbatim}
2
2
3
3
3
\end{verbatim}
and therefore $R=3$.

Given that our set $\mathcal{E}$ (or $\overline{\mathcal{E}}$) covers all combinations of results, we are choosing to prioritize those cases where our different algorithm-parameter-metric results are cohesive (row-wise), before looking at their actual value (column-wise).

We present results for the three approaches in Section 4.

\section{Results}
\label{sec:4}

Now, we arrive at our results. In the following sections, we present the results of the 6 different approaches (Raw/Mode with Row/Column/Full) on 100 simulated datasets from Section~\ref{sec:21}. We compare our results to 2 benchmarks.

The first is the expected value from randomly sampling our result set 100 times - essentially the accuracy we could expect from choosing an algorithm, hyperparameters and metrics beforehand.

The second is the Consensus Clustering approach put forward by Monti et al~\cite{monti}. In this case, we used K-means and GMM and attempted to pass in both default hyperparameters (D) and the best performing hyperparameters (B) as determined by our method.

\subsection{Performance}

Here, we define accuracy by comparing the predicted number of clusters for each of the 100 datasets - based on voting on the $\mathcal{E}$ or $\overline{\mathcal{E}}$ set - to the actual number of clusters (3 in every case).

Overall results are shown in Table 4.

\begin{table}[h!]
	\caption{Accuracy on 100 simulated datasets.}
	\centering\begin{tabular}{r|rr}
		
		& Raw &  Mode \\
		
		Full &  86.00 &  \textbf{93.00} \\
		Row &  89.00 &  \textbf{92.00} \\
		Col &  88.00 &  \textbf{90.00} \\
		
	\end{tabular}
\end{table}

While there are differences in performance between the voting methods, the signal is somewhat muddled. In the case of the Raw construction the Row-first approach is best, while for a Mode construction a Full vote is preferable. Overall, the differences in voting performance is also small providing at most a $3\%$ increase. We don't consider it prudent to declare one voting approach more beneficial than another.

On the other hand, the choice between using a Raw ensemble construction $\mathcal{E}$ or a mode-based $\overline{\mathcal{E}}$ set seems to be clearer. Using a mode construction improved performance across the board, yielding a $2-8\%$ increase in accuracy.

Given a mode-based $\overline{\mathcal{E}}$ set, the naive voting takes the lead - it is overall the best performer with $93\%$ accuracy. Additionally, we find it important to note that we have actually outperformed even the maximum performance we saw in Table 1 - which was $91\%$.

Given that the latter could only occur given a prefect guess as to which algorithm-hyperparameter-metric combination to use, we find it even more satisfying.

\subsection{Benchmarks}

Table 5 details our benchmark performance as defined in this section - Consensus Clustering and expected value from random sampling.
\begin{table}[h!]
	\caption{Benchmark accuracy on 100 simulated datasets.}
	\centering\begin{tabular}{r|rr}
		
		& Expected Value &  Consensus \\
		
		--- &  \textbf{57.29} &  --- \\
		KMeans (D) &  --- &  82.00 \\
		KMeans (B) &  --- &  \textbf{87.00} \\
		GMM (D) &  --- &  66.00 \\
		GMM (B) &  --- &  46.00 \\
		
	\end{tabular}
\end{table}

Not unexpectedly, randomly guessing at possible solutions yields unsatisfactory results. We note that it consistently scores above $50\%$, likely due to the fact that even poor algorithm configurations can still pick up some signal.

Consensus Clustering gave very unpredictable results, with a very large variance in performance - though it did peak at a respectable $87\%$. We reiterate our previous point, however, that it requires a choice of algorithm and hyperparameters as inputs - greatly reducing its effectiveness in practice.

Perhaps the most interesting result to come out of benchmarking was the performance of Consensus Clustering with respect to hyperparameter selections that did very well in our method. In the case of GMM this lead to a drastic drop in performance ($30\%$), while for K-means we saw a $6\%$ increase. This would indicate that while it is likely a non-trivial exercise to combine the approaches in a reasonable way, it could be worth further investigation.

\section{Usage}
\label{sec:5}
Now that we have examined this problem and our proposed solution, we'd like to discuss some other elements. Namely, typical usage setups, and some simple approaches to selecting the best algorithm-hyperparameter combination in each case. As our colleagues in industry would say - how do we use this in production?

We would like to note that the  following algorithm selection methods in particular are merely simple approaches to get things off the ground. There are undoubtedly other ways to solve this and we encourage further work in this area.

We believe there to be two general use-cases depending mainly on computational constraints: the case where all our data can be processed at once, and the case where we must slice our data into subsets first.

\subsection{Data Subsets}

If we look at the case where we must partition our data due to computational limitations, we find ourselves in essentially the same framework that we had throughout the paper. While we will leave it to the reader to determine the best way to sample subsets of their data while capturing all clusters, let's examine how this ensemble framework would work.

Throughout, we have looked at $100$ simulated datasets as a means of getting accuracy metrics. Suppose now we take our large dataset and split it into $100$ smaller, more manageable, datasets. We are now in the same situation as we were earlier in the paper: we would expect the number of clusters to be the same for each of the smaller datasets, and we would aggregate the results of $100$ ensemble methods (albeit not with ``accuracy").

\subsubsection{Estimating the Number of Clusters}

Now that we have our $100$ subsets, we can construct our $E$ matrix for each one of them and vote on the number of clusters. This will give us $100$ answers, one cluster number estimate per subset.

From there, we could select the most common answer (ie, the mode) as our global estimated number of clusters (instead of computing accuracy as we did in this paper). Having determined how many clusters our dataset has, we arrive at the question of choosing an algorithm.

\subsubsection{Algorithm-Hyperparameter Selection}

Still within our $100$ subset context, let's examine how we could choose a best algorithm-hyperparameter combination. We reiterate that this only one of what is likely many possible approaches.

Given our estimated number of clusters, store all the answers estimated by each algorithm-hyperparameter-metric combination found in the subset $E$ matrices. From there, compute the accuracy of each algorithm-hyperparameter-metric combination (relative to our global estimated number of clusters). The best-performing combination can be taken as a reasonable way of clustering future data from the same dataset.

Indeed, in our simulated case, this approach correctly identifies that $GMM$ is the best choice, and more specifically that the combination
\begin{verbatim}
algorithm:
    GMM
hyperparameters:
    covariance_type: diag,
    reg_covar: 1e-8
metric:
    AIC
\end{verbatim}
achieves the best results, with $91\%$ accuracy. Readers will note this is indeed the top performance we can achieve as per Table 1.

It seems reasonable to assume that given more data drawn from the same dataset, this algorithm with these hyperparameters would do well at clustering it in a sensible way. We do note that it is possible to apply the logic presented in Section 5.2.2 in this case as well, such logic is included in the code library by default.

\subsection{Full Dataset}

Now let's look at the simpler case where all of our data can be processed together as a single dataset.

\subsubsection{Estimating the Number of Clusters}

In this happy scenario, estimating the number of clusters is relatively straightforward - we run a single workflow. We begin be constructing our $E$ matrix. Then, we compute the results of voting - which directly gives us the predicted number of clusters in our data.

In this case there is no need for any aggregation as we have a single outcome from the vote.

\subsubsection{Algorithm-Hyperparameter Selection}

When it comes to identifying the right choice of combination, however, we can't proceed as we did in the subset case. Given that we have no way of computing accuracy, we will instead look for the ``most stable" choice. Once more, this is simply a first approach and that future work could likely result in improvements.

For each algorithm-hyperparameter combination, look at its predictions across metrics (which are not needed for future clustering). For example, suppose that on the first dataset from our simulated data the $GMM$ combination
\begin{verbatim}
algorithm:
    GMM
hyperparameters:
    covariance_type: diag,
    reg_covar: 1e-8
\end{verbatim}
obtained results of \verb|[3, 3, 3, 3]| across its four metrics, whereas
\begin{verbatim}
algorithm:
    GMM
hyperparameters:
    covariance_type: spherical,
    reg_covar: 1e-2
\end{verbatim}
obtained \verb|[3, 4, 2, 3]|.

In this example, we would favor the combination that was most often correct: \verb|[3, 3, 3, 3]|. Note that we are essentially looking for stability and insensitivity to metric choice.

If we extend this comparison to every algorithm-hyperparameter combination, we arrive at a suitable combination choice to use for future clustering. We note, however, that given the small number of metric choices, this approach is less likely to yield a unique best choice.

If that is the case, any of the top-ranked combinations may be selected, as they are equally likely to achieve desirable results - as per this stability framework.

\section{Conclusion}
\label{sec:6}

We have developed a workflow with 6 possible configurations for determining the number of clusters in an unlabeled dataset, while offering a flexible basis for determining the optimal choice of algorithm and  associated hyperparameters. While certain methods already exist, such as for agglomerative hierarchical clustering and Consensus Clustering, they each present difficulties in the field that our approach addresses.

Fistly, we no longer require researchers - who may not be subject matter experts in unsupervised learning, but rather their own domains - to provide such specific inputs as particular algorithms and hyperparameter-metric configurations. Instead, given reasonably spanning ranges of hyperparameters, and a diverse selection of algorithms, we can reliably predict the number of clusters present with more than $90\%$ accuracy.

We also obtain - at very little cost - a reasonably suitable choice for which algorithm and which hyperparameters to use to further cluster the data. While not every situation will require clustering of additional incoming data from the same distribution, combination performance findings are sure to be beneficial to researchers.

Lastly, it is our hope that the simple algorithmic structure of this approach can lead to reliably simple integration into commonly used software libraries - thereby removing another barrier to entry. The code used for this framework can be found at~\textsc{https://github.com/antoinezambelli/ensemble-clustering}.

In the future we hope to explore several avenues of work related to this approach. This includes studying other means of measuring cohesion (other than a row-first approach). We would also like a process to automatically track the ensemble set as it is populated and to adjust the hyperparameter space dynamically. This could allow for faster computations and less noise in the resulting ensemble. Finally, a careful and constructive combination of Consensus Clustering and our ensemble method.

\section*{Acknowledgment}

We thank Dr. Alexandra Cunliffe for her help in reorganizing the paper for better clarity.

\end{document}